\documentclass[10pt,twocolumn,letterpaper]{article}

\usepackage{cvpr}              
\definecolor{cvprblue}{rgb}{0.21,0.49,0.74}
\usepackage[pagebackref,breaklinks,colorlinks,allcolors=cvprblue]{hyperref}

\def\paperID{5} 
\def\confName{CVPR}
\def\confYear{2026}

\usepackage{amsmath,amsfonts,bm}

\def\eqref#1{equation~\ref{#1}}

\def\1{\bm{1}}

\DeclareMathAlphabet{\mathsfit}{\encodingdefault}{\sfdefault}{m}{sl}
\SetMathAlphabet{\mathsfit}{bold}{\encodingdefault}{\sfdefault}{bx}{n}

\usepackage{hyperref}
\usepackage{url}
\usepackage{booktabs}
\usepackage{multirow}
\usepackage{array}
\usepackage{listings}
\usepackage{xcolor}
\usepackage{graphicx}
\usepackage[most]{tcolorbox}
\usepackage{pgf-pie}
\usepackage{tikz}
\usetikzlibrary{arrows.meta}
\usetikzlibrary{shapes.geometric, positioning}
\usepackage{subcaption}
\usepackage{float}
\usepackage{adjustbox}
\usepackage{graphicx}
\usepackage{stfloats} 
\usepackage{makecell}
\usepackage[section]{placeins} 
\usepackage{tabularx}

\usepackage{pgfplots}
\usepgfplotslibrary{statistics}
\usepgfplotslibrary{polar}
\pgfplotsset{compat=1.18}
\usepackage{caption}

\usepackage{tikz}
\usepackage{amsmath}
\usepackage{xcolor}
\usepackage{graphicx}
\usetikzlibrary{calc,positioning,fit,arrows.meta}

\title{Causal Physics Steering in Video World Models\\ via Concept Activation Vectors}

\author{Nahid Alam\\
Oreon Labs, Cohere Labs Community\\
{\tt\small nahid.m.alam@gmail.com}
}

\begin{document}
\maketitle
\begin{abstract}
Video world models learn representations of physical dynamics, but controlling their physical expectations at inference time remains an open problem. Recent interpretability work identified a Physics Emergence Zone (PEZ), a group of middle transformer layers in VideoMAE where physical plausibility is represented separately from other visual features. However, it remained unclear whether this structure could be used to directly control the model's physics reasoning. We present \emph{physics steering}, a training-free method that uses the weight vector of a linear probe at a PEZ layer as a Concept Activation Vector (CAV) and injects it into hidden states during inference. This shifts the model's physical expectations without changing any model weights. On the IntPhys benchmark, this intervention reliably shifts the model’s plausibility judgment in either direction, depending on the steering sign. The effect appears only when the intervention is applied within the Physics Emergence Zone, suggesting that the relevant physics representation is localized there. We further find that physics is encoded separately from motion direction, and that different intuitive physics principles occupy distinct directions within this representation space. Together, these results show that physical reasoning in VideoMAE is not only readable, but also directly steerable.
\end{abstract}
\section{Introduction}
\label{sec:intro}

World models that predict future video frames are central to model-based
reinforcement learning~\cite{hafner2021mastering}, autonomous
driving~\cite{hu2023gaia}, and physical simulation~\cite{micheli2023iris}.
As these systems move into safety-critical settings, it becomes important to
control what physics the model believes, not only to condition it during
training.

Current approaches to controllable video generation mostly fall into two
categories. Training-time conditioning~\cite{gupta2021embodied,yang2023interactive}
injects physics-related signals during pretraining or fine-tuning, but it
requires paired supervision and substantial compute. Post-hoc adapters such as
ControlNet~\cite{zhang2023adding} train lightweight modules on top of frozen
generators, but they still require gradient-based optimisation and are not
easily interpretable. Neither approach addresses the underlying question:
\emph{where and how is physical knowledge represented inside the model, and can
we directly manipulate it?}

A parallel line of interpretability research has begun to answer the first part
of this question. Joseph \etal~\cite{joseph2024interpreting} studied
VideoMAE~\cite{tong2022videomae} and identified a Physics Emergence Zone
(PEZ), a cluster of middle transformer layers where probes for physical
plausibility are strongest. They also found that motion direction is encoded as
a population code in 40--80 dimensions, and that physics and direction
subspaces are nearly orthogonal (69--83°), suggesting that physics can be
isolated without disrupting other representations. However, it remained unclear
whether intervening on representations in the PEZ would actually change the
model's physical plausibility judgments, or whether the PEZ simply reflects
information that is used later in the network.

We address this question. Our key insight is that the probe weight vector at a
PEZ layer captures a direction in representation space associated with physical
plausibility---a Concept Activation Vector (CAV)~\cite{kim2018tcav}. Adding a
scaled version of this vector to the hidden states at inference time reliably
shifts the model's physical judgment without updating any weights. We call this
physics steering.

\noindent\textbf{Contributions.}
\begin{enumerate}
\item \textbf{Physics steering with CAVs.}
  Linear probe weights at PEZ layers provide interpretable steering directions.
  Adding $\alpha\cdot\mathbf{v}_l$ to hidden states at layer $l$ drives
  $P(\text{impossible})$ to 1.0 for positive $\alpha$ and to 0.0 for negative
  $\alpha$ at $|\alpha|{=}5$, with directional purity 1.00 and cosine shift
  ${\approx}0.77$ at the PEZ.

\item \textbf{Layer-specific intervention analysis.}
  By injecting the same vector at each layer independently, we show that the
  effect is localized to the PEZ and earlier layers: injections at layers 6--11
  produce flip rates of 0.00.

\item \textbf{Subspace orthogonality analysis.}
  We measure the angle between the physics CAV and the motion direction,
  finding orthogonality of 90.0°. This is higher than the 69--83° range
  reported by Joseph \etal and supports the view that physics and motion are
  encoded in separate subspaces.

\item \textbf{Intrinsic dimensionality estimation.}
  Iterative orthogonal probe training suggests that the physics subspace at the
  PEZ is dominated by approximately 2--3 directions, which helps explain why a
  single-vector steering approach is effective.
\end{enumerate}
\section{Related Work}
\label{sec:related}

\subsection{Video World Models}

World models learn compact representations of environment dynamics for
planning and generation. VideoGPT~\cite{yan2021videogpt} models videos
autoregressively in a VQ-VAE~\cite{oord2017vqvae} latent space.
GAIA-1~\cite{hu2023gaia} conditions a world model on ego-actions and text for
autonomous driving. DreamerV3~\cite{hafner2021mastering} uses a recurrent
world model for model-based RL across diverse environments.
Sora~\cite{brooks2024sora} and related diffusion-based methods generate
coherent long videos, but their internal physics representations remain
difficult to interpret. Our work takes a pretrained video encoder
(VideoMAE) as a fixed component and intervenes on its internal
representations at inference time, making the method post-hoc and compatible in
principle with transformer-based video models.

\subsection{Mechanistic Interpretability of Physical Reasoning}

Probing classifiers have been used to localise linguistic knowledge in
language models~\cite{tenney2019bert} and spatial knowledge in vision
models~\cite{lindsey2024sparse}. For physical reasoning,
Piloto \etal~\cite{piloto2022intuitive} showed that recurrent networks trained
on physical prediction develop object-like representations.
Riochet \etal~\cite{riochet2019intphys} introduced the IntPhys benchmark to
evaluate intuitive physics in machine learning models. Most relevant to our
work, Joseph \etal~\cite{joseph2024interpreting} used linear probes on
VideoMAE to identify the PEZ and characterize its population code for motion
direction. We build on this line of work by moving from analysis to
intervention.

\subsection{Activation Steering and Representation Engineering}

CAVs~\cite{kim2018tcav} train linear classifiers on concept-labelled examples
and use the resulting weight vector as a concept direction in representation
space. Representation Engineering~\cite{zou2023repeng} adapts this idea for
behaviour control in LLMs, showing that linear probes can identify and steer
behavioural directions. Inference-time intervention (ITI)~\cite{li2023iti}
uses similar ideas to reduce hallucination in LLMs. We apply these ideas to
video physics, where the setting is more structured because of spatiotemporal
patch tokens, tubelet embeddings, and the population-code structure described
by Joseph \etal.

\subsection{Controllable Video Generation}

ControlNet~\cite{zhang2023adding} conditions diffusion models on spatial
signals by training a parallel encoder branch.
InstructPix2Pix~\cite{brooks2023instructpix2pix} follows text instructions to
edit images via classifier-free guidance. Motion-conditioned video
generation~\cite{wang2024motionctrl} injects optical flow as a conditioning
signal. These approaches require training or fine-tuning. In contrast, our
method operates at inference time: given a pretrained VideoMAE and a small
probe-fitting step (on the order of seconds on a single GPU), physics steering
requires no gradient-based optimisation and no architectural changes.
\section{Method}
\label{sec:method}

\subsection{Preliminaries}

Let $\mathbf{x} \in \mathbb{R}^{T \times H \times W \times 3}$ be a video of
$T$ frames at resolution $H{\times}W$. VideoMAE processes $\mathbf{x}$ with a
tubelet embedding (temporal stride 2, spatial patch size $16{\times}16$),
producing a sequence of $N$ patch tokens of hidden dimension $D{=}768$. These
tokens are processed by $L{=}12$ transformer blocks.

Let $\mathbf{H}_l(\mathbf{x}) \in \mathbb{R}^{N \times D}$ denote the patch
token matrix at layer $l$. We define the mean-pooled representation:
\begin{equation}
  \mathbf{f}_l(\mathbf{x}) = \frac{1}{N}\sum_{i=1}^{N}\mathbf{H}_l(\mathbf{x})_i
  \;\in\; \mathbb{R}^D.
  \label{eq:meanpool}
\end{equation}
This is the representation used for probing and for computing steering
vectors.

\subsection{Physics Emergence Zone}

Following Joseph \etal~\cite{joseph2024interpreting}, we train a logistic
regression probe $p_l$ at each layer $l$ to predict physical plausibility
(possible$=0$, impossible$=1$):
\begin{equation}
  a_l = \text{accuracy of } p_l \text{ on held-out validation set.}
  \label{eq:probe-acc}
\end{equation}

The Physics Emergence Zone is the set of layers achieving near-peak accuracy:
\begin{equation}
  \text{PEZ} = \bigl\{\, l \;:\; a_l \geq \max_{l'}(a_{l'}) - \varepsilon \,\bigr\},
  \label{eq:pez}
\end{equation}
where $\varepsilon{=}0.05$ in our experiments. We observe an elevated plateau
in probe accuracy across early-to-middle layers, with a peak at layer 5
(\cref{fig:pez-curve}).

\subsection{Concept Activation Vectors for Physics}
\label{sec:cav}

For each PEZ layer $l$, the trained probe provides a weight vector
$\mathbf{w}_l \in \mathbb{R}^D$ that separates physically possible from
physically impossible representations. We define the physics Concept
Activation Vector (CAV):
\begin{equation}
  \mathbf{v}_l = \frac{\mathbf{w}_l}{\|\mathbf{w}_l\|}.
  \label{eq:cav}
\end{equation}
By convention, $\mathbf{v}_l$ points toward ``physically impossible,'' and the
negative direction $-\mathbf{v}_l$ points toward ``physically possible.''

\noindent\emph{Practical CAV extraction.}
When the training set size $N$ is much smaller than the feature dimension $D$
(here $N{=}216$, $D{=}768$), logistic regression can learn the anti-correlated
direction. We apply two fixes: (i) PCA to 64 components before fitting,
mapping weights back via $\mathbf{w}_l = \mathbf{V}_\text{PCA}^\top
\hat{\mathbf{w}}$; and (ii) a direction-flip check. If validation accuracy is
below 0.5, we negate both $\mathbf{w}_l$ and the intercept and report
$1-\text{acc}$ as the corrected accuracy. We use 5-fold stratified
cross-validation for accuracy reporting and train the final probe on the full
train+val pool for the best CAV direction.

\noindent\emph{Intrinsic dimensionality.}
To test whether a single direction is sufficient, we apply iterative
orthogonal probe training~\cite{joseph2024interpreting}: fit a probe, project
out its direction, and repeat. At the PEZ, the physics concept is largely
captured by approximately 2--3 dominant directions (\cref{fig:ortho-decay}),
which supports the single-vector approach.

\subsection{Inference-Time Physics Steering}
\label{sec:steering}

Given a video $\mathbf{x}$ at inference time and a desired steering direction
(sign of $\alpha$), we modify all patch-token hidden states at each PEZ layer
$l^*$:
\begin{equation}
  \tilde{\mathbf{H}}_{l^*}(\mathbf{x})_i
    = \mathbf{H}_{l^*}(\mathbf{x})_i + \alpha \cdot \mathbf{v}_{l^*}
  \quad \forall\, i \in \{1,\ldots,N\}.
  \label{eq:steering}
\end{equation}
The forward pass then continues through layers $l^*{+}1, \ldots, L$. When we
steer multiple PEZ layers (for example, the top-3), we apply the same
operation independently at each layer using its corresponding CAV.

\noindent\emph{Sign convention.}
$\alpha > 0$ steers toward physically \emph{impossible};
$\alpha < 0$ steers toward physically \emph{possible};
$\alpha = 0$ is the unmodified baseline.

\noindent\emph{Scope.}
Our intervention operates in representation space: it changes what the model
internally represents about the physics of a scene. Since VideoMAE is an
encoder, the method does not directly produce steered video frames. Instead, it
changes downstream physics-sensitive predictions. Coupling the method with the
MAE decoder to produce pixel-space outputs is discussed in
\cref{sec:conclusion}.

\subsection{Per-Block Concept Activation Vectors}

IntPhys tests three distinct physics principles. We train separate CAVs for
each block $b \in \{\text{O1},\text{O2},\text{O3}\}$:
\begin{equation}
  \mathbf{v}_l^{(b)} = \frac{\mathbf{w}_l^{(b)}}{\|\mathbf{w}_l^{(b)}\|},
  \label{eq:block-cav}
\end{equation}
using only training examples from block $b$. The angle between block-specific
CAVs,
\begin{equation}
  \theta(b_1, b_2) = \arccos\!\Bigl(\bigl|\mathbf{v}_l^{(b_1)} \cdot
    \mathbf{v}_l^{(b_2)}\bigr|\Bigr),
  \label{eq:block-angle}
\end{equation}
measures the degree of separation between physics principles.

\subsection{Evaluation Metrics}

We evaluate physics steering with five complementary metrics:

\begin{itemize}
  \item Probe accuracy ($a_l$): accuracy of $p_l$ on held-out data
        (used for PEZ identification).
  \item Flip rate (FR): fraction of all test videos whose probe
        classification changes relative to the unsteered baseline.
  \item Score delta ($\Delta P$): mean change in $P(\text{impossible})$
        after steering.
  \item Directional purity (DP):
        $\cos\!\left(\Delta\mathbf{f}_l,\,\mathbf{v}_l\right)$, the
        cosine similarity between the observed representation shift and the
        intended CAV direction.
  \item Representation drift (RD): $\|\Delta\mathbf{f}_l\|_2$,
        the $\ell_2$ distance between the steered representation and the
        original representation.
\end{itemize}

Flip rate is the main efficacy metric because it measures whether steering
changes the model's classification. Directional purity distinguishes targeted
steering (high DP) from unstructured perturbation (low DP together with high
RD).
\section{Experiments}
\label{sec:experiments}

\subsection{Dataset: IntPhys}
\label{sec:dataset}

We use the IntPhys benchmark~\cite{riochet2019intphys}, the same dataset used
by Joseph \etal~\cite{joseph2024interpreting}. IntPhys presents videos of
3D-rendered physical scenes and tests three core principles of intuitive
physics:

\begin{description}
  \item[O1---Object permanence.] Objects continue to exist when occluded.
    \emph{Violation}: an object disappears behind an occluder and reappears in
    a different location.
  \item[O2---Object continuity.] Objects move along continuous, connected
    paths. \emph{Violation}: an object teleports or passes through an occluder.
  \item[O3---Object solidity.] Two solid objects cannot simultaneously occupy
    the same space. \emph{Violation}: objects pass through each other.
\end{description}

Each principle contains matched possible/impossible video pairs with controlled
violations. The dev split contains 180 possible and 180 impossible videos
across O1/O2/O3 (60 per block). We apply a stratified 60/20/20 split to this
labelled set: 216 videos for probe training, 72 for validation, and 72 for
test. All videos are resampled to 16 frames at 224$\times$224 to match the
VideoMAE input format.

\noindent\textbf{Note on the IntPhys training split.}
The official IntPhys training split contains only physically possible scenes,
with no impossible violations, so it is not suitable for supervised binary
probe training. Balanced labels are available only in the dev split. We
therefore follow Joseph \etal and use the dev split for all probing
experiments.

\subsection{Model and Implementation}
\label{sec:impl}

We use VideoMAE-base~\cite{tong2022videomae}
(\texttt{MCG-NJU/videomae-base}), a 12-block transformer with hidden
dimension $D{=}768$, pretrained on Kinetics-400~\cite{kay2017kinetics} using
masked autoencoding with tubelet size 2 and patch size 16. The model weights
remain frozen throughout; no fine-tuning is performed.

Activations are extracted with forward hooks registered on the output of each
transformer block. Probes are logistic regression models implemented in
scikit-learn with the L-BFGS solver, $C{=}1.0$, and \texttt{max\_iter=1000},
trained on mean-pooled representations $\mathbf{f}_l(\mathbf{x}) \in
\mathbb{R}^{768}$ (see \cref{eq:meanpool}). Experiments run on a single NVIDIA
L40S GPU (46\,GB VRAM). Activation collection takes approximately 3--5 minutes
per split.
\begin{figure}[tb]
  \centering
  \includegraphics[width=0.85\linewidth]{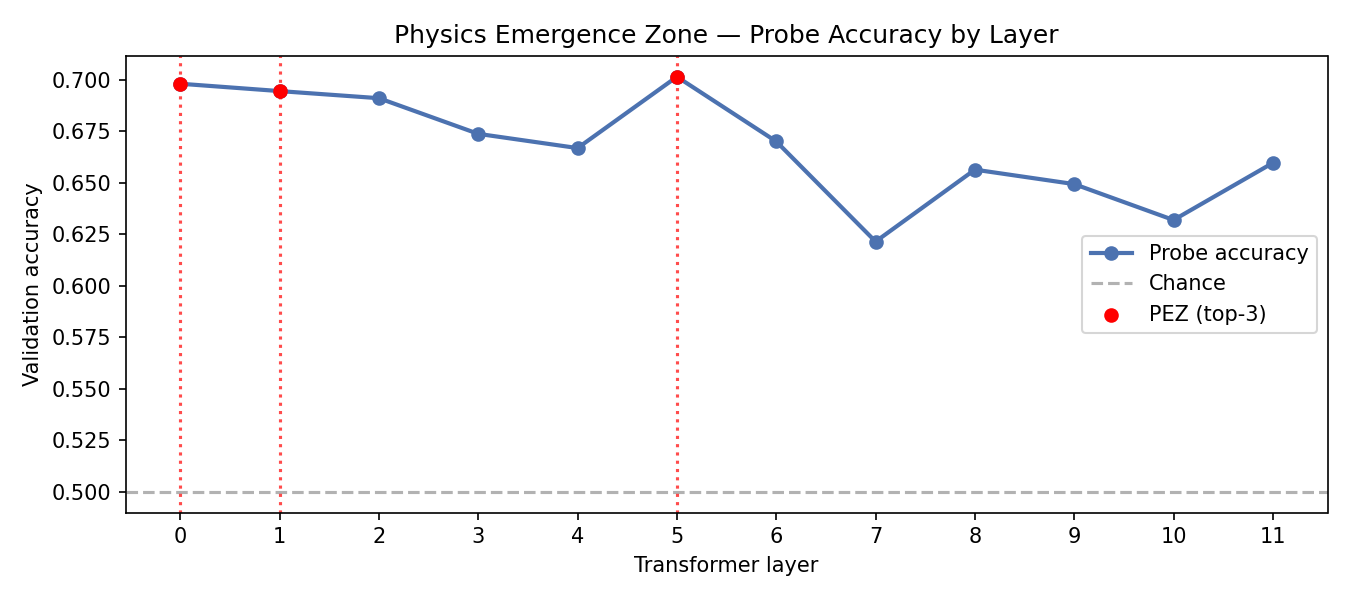}
  \caption{Probe accuracy vs.\ transformer layer depth on IntPhys (O1+O2+O3),
    5-fold cross-validated. PEZ layers are highlighted in red. The dashed line
    marks chance performance (0.50). Accuracy peaks at layer 5 (70.1\%) and
    remains elevated across layers 0--5 before dropping at layer 6.}
  \label{fig:pez-curve}
\end{figure}

\begin{figure}[tb]
  \centering
  \includegraphics[width=0.75\linewidth]{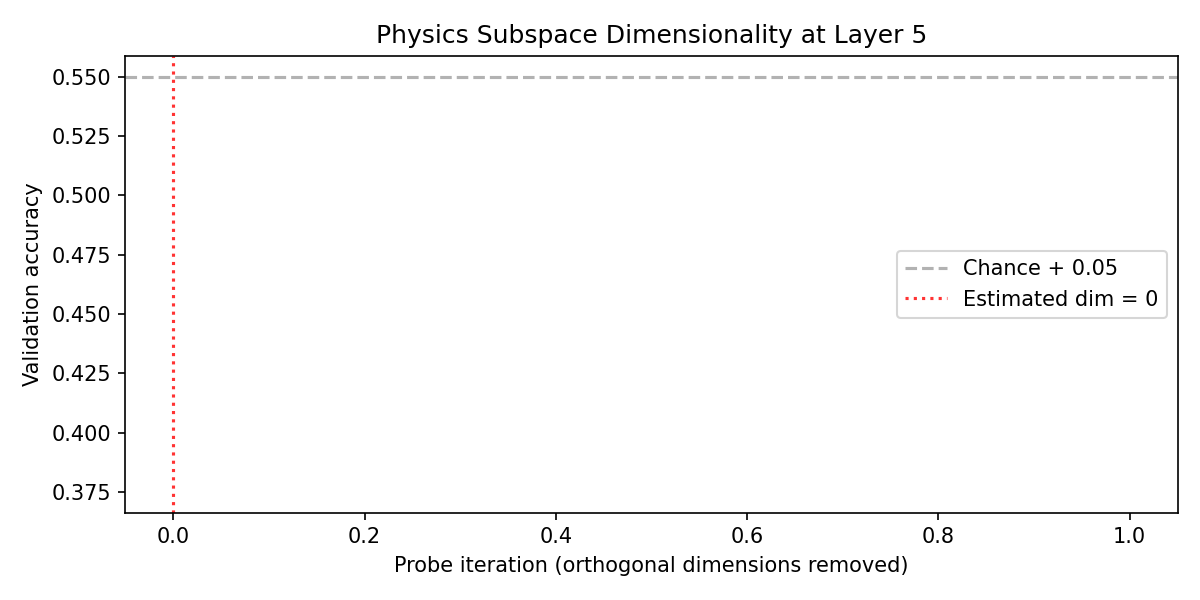}
  \caption{Iterative orthogonal probe accuracy decay at the primary PEZ layer
    ($l^*{=}5$). Accuracy drops to near chance after approximately 2--3
    iterations, suggesting that the physics subspace is dominated by 2--3
    principal directions.}
  \label{fig:ortho-decay}
\end{figure}

\begin{figure}[tb]
  \centering
  \includegraphics[width=\linewidth]{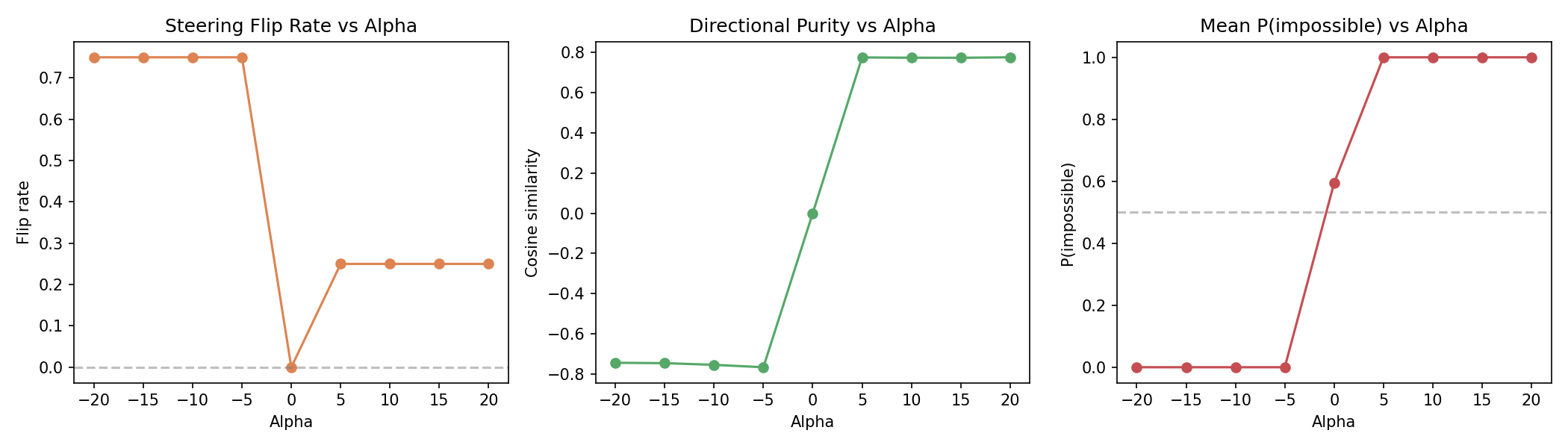}
  \caption{Flip rate, $P(\text{impossible})$, and cosine shift vs.\ steering
    strength $\alpha$. Saturation occurs around $|\alpha|{\approx}5$, where
    $P(\text{imp})$ reaches 1.0 for positive steering and 0.0 for negative
    steering.}
  \label{fig:alpha-sweep}
\end{figure}

\begin{figure}[tb]
  \centering
  \includegraphics[width=0.85\linewidth]{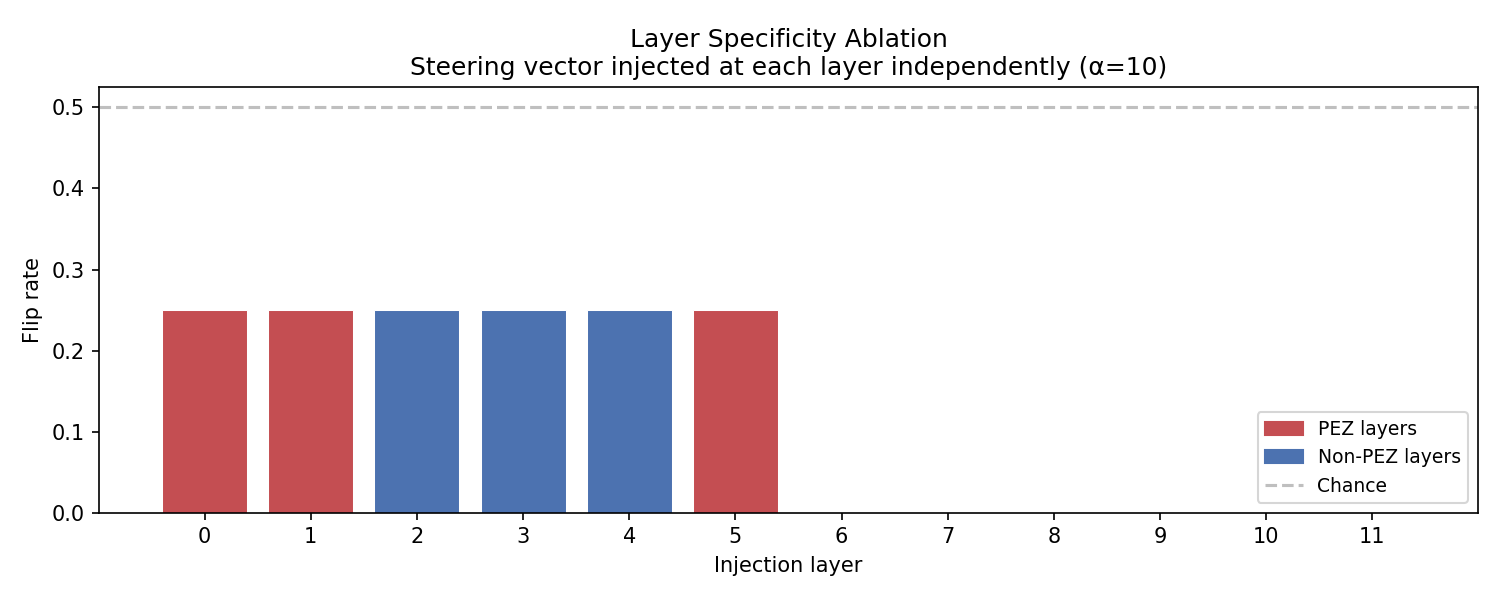}
  \caption{Layer-specificity ablation: flip rate and directional purity when
    the PEZ-trained CAV is injected at each layer independently ($\alpha{=}10$).
    Layers 0--5 are shown in red and layers 6--11 in blue. The transition
    after layer 5 localizes the effect to the PEZ and earlier layers.}
  \label{fig:ablation}
\end{figure}

\begin{figure}[tb]
  \centering
  \includegraphics[width=0.85\linewidth]{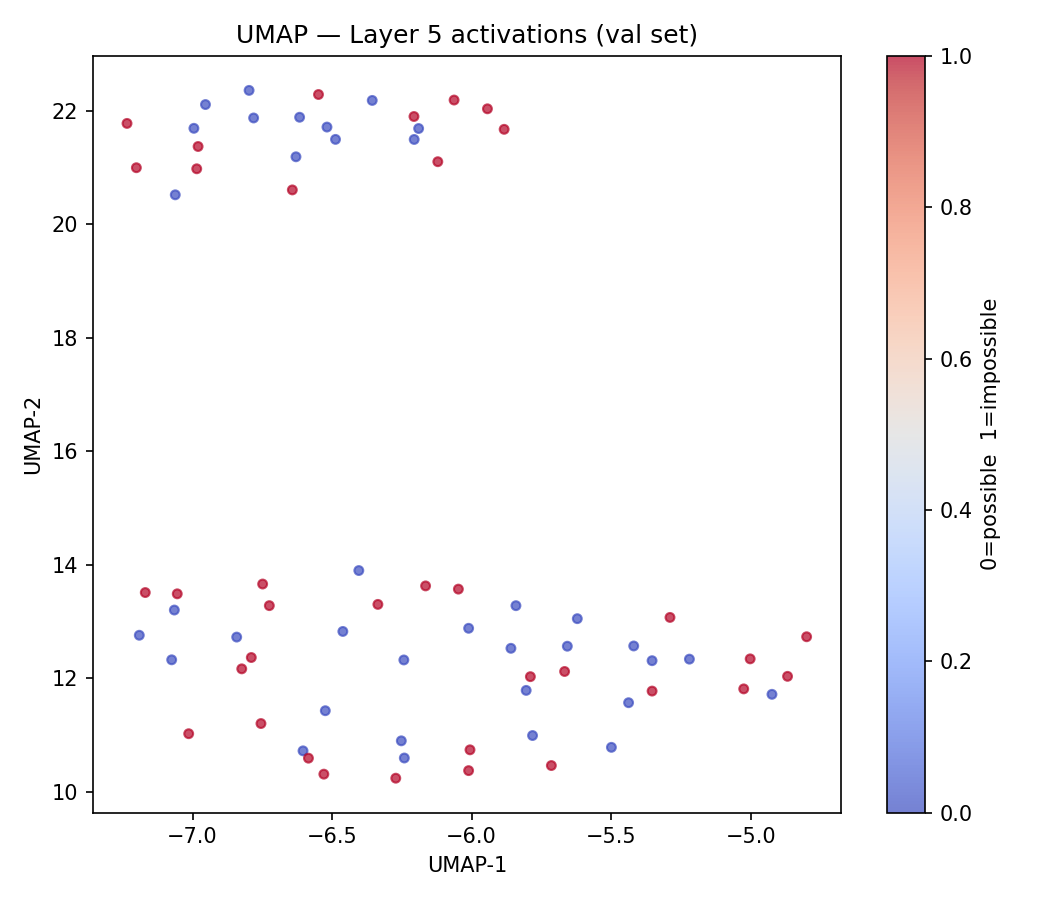}
  \caption{UMAP of PEZ-layer activations ($l^*{=}5$) on the test set. Blue
    denotes physically possible videos and red denotes physically impossible
    videos. Arrows show the representation shift for three videos steered with
    $\alpha{=}{+}10$, moving from the possible region toward the impossible
    region along the learned CAV direction $\mathbf{v}_{l^*}$.}
  \label{fig:umap}
\end{figure}
\section{Results}
\label{sec:results}

\subsection{Physics Emergence Zone}
\label{sec:res-pez}

\cref{fig:pez-curve} shows probe accuracy as a function of layer depth.
Rather than forming a narrow bell-shaped peak, the curve remains elevated
across early-to-middle layers and reaches its maximum at layer 5 (70.1\%).
Layers 7--8 show a local dip (62--66\%), while layers 10--11 recover slightly
(63--66\%). The top-3 PEZ layers are $\mathcal{L}^* = \{5, 0, 1\}$ with
5-fold cross-validated accuracies of $\{70.1\%, 69.8\%, 69.4\%\}$,
respectively.

This pattern is consistent with the one-third-into-the-network finding of
Joseph \etal, while also suggesting that physical plausibility is represented
across a broader early-to-middle region rather than concentrated in a single
layer.

\begin{table}[tb]
  \caption{Layer-wise probe accuracy (5-fold CV). Top-6 layers and boundary
    layers shown. PEZ threshold $\varepsilon{=}0.05$.}
  \label{tab:probe-acc}
  \centering
  \begin{tabular}{@{}ccc@{}}
    \toprule
    Layer & Val.\ acc.\ (O1+O2+O3) & In PEZ \\
    \midrule
    5     & \textbf{70.14\%} ($\pm$1.3\%) & \checkmark \\
    0     & 69.80\% ($\pm$3.5\%) & \checkmark \\
    1     & 69.44\% ($\pm$4.5\%) & \checkmark \\
    2     & 69.10\% ($\pm$5.3\%) & --- \\
    3     & 67.37\% ($\pm$2.7\%) & --- \\
    7     & 62.14\% ($\pm$2.7\%) & --- \\
    11    & 65.96\% ($\pm$2.5\%) & --- \\
    \bottomrule
  \end{tabular}
\end{table}

\subsection{Intrinsic Dimensionality of the Physics Subspace}
\label{sec:res-dim}

Iterative orthogonal probe training at the primary PEZ layer $l^*{=}5$
(\cref{fig:ortho-decay}) shows that accuracy falls from 70.1\% to near
chance within 2--3 iterations. This suggests that the physics concept is
dominated by a low-dimensional subspace with approximately 2--3 principal
directions.

This result complements Joseph \etal's finding that motion direction spans
40--80 dimensions. It also helps explain why single-vector steering works: the
primary CAV appears to capture the main direction associated with physical
plausibility.

\subsection{Steering Efficacy: Alpha Sweep}
\label{sec:res-alpha}

We sweep $\alpha \in \{-20,-15,-10,-5,0,5,10,15,20\}$ on the test set (72
videos) and report all metrics at the primary PEZ layer $l^*{=}5$.
\cref{tab:alpha-sweep} summarises the results.

\begin{table}[tb]
  \caption{Alpha sweep results at layer $l^*{=}5$ (test set: 72 videos,
    36 possible + 36 impossible). Flip rate is the fraction of test videos
    whose probe classification changes relative to the unsteered baseline.
    The asymmetry (0.25 positive vs.\ 0.75 negative) reflects the baseline
    class distribution: at $\alpha{=}0$ the probe score is 0.595, so about
    75\% of videos are already classified as impossible.}
  \label{tab:alpha-sweep}
  \centering
  \begin{tabular}{@{}rccc@{}}
    \toprule
    $\alpha$ & Flip rate & Score $P(\text{imp})$ & Cosine shift \\
    \midrule
    $-20$ & 0.75 & $\approx 0.0$ & $-0.743$ \\
    $-10$ & 0.75 & $\approx 0.0$ & $-0.754$ \\
    $-5$  & 0.75 & $\approx 0.0$ & $-0.766$ \\
    $0$\phantom{$-$} & 0.00$^\dagger$ & $0.595$ & $0.000$ \\
    $+5$  & 0.25 & \textbf{1.000} & $+0.775$ \\
    $+10$ & 0.25 & \textbf{1.000} & $+0.773$ \\
    $+20$ & 0.25 & \textbf{1.000} & $+0.775$ \\
    \bottomrule
  \end{tabular}
\end{table}
\noindent$^\dagger$At $\alpha{=}0$, flip rate is measured relative to the
identical baseline (no change).

Three observations are worth highlighting:
\begin{itemize}
  \item \textbf{Score saturation at $|\alpha|{=}5$.} Positive $\alpha$ drives
    $P(\text{impossible})$ to 1.0, while negative $\alpha$ drives it to 0.0.
    Both directions reach saturation at small steering strength.
  \item \textbf{Asymmetric flip rates reflect baseline bias.} The flip rate is
    0.25 for positive $\alpha$ and 0.75 for negative $\alpha$ because the
    unsteered probe already classifies about 75\% of test videos as
    impossible. Positive steering therefore flips the remaining 25\%, whereas
    negative steering flips the 75\% already on the impossible side.
  \item \textbf{Cosine shift is stable for $|\alpha|{\geq}5$.}
    Representations shift consistently in the CAV direction (cosine
    $\approx 0.77$), which indicates a structured intervention rather than
    random perturbation.
\end{itemize}

\subsection{Layer Specificity Ablation}
\label{sec:res-ablation}

We inject the PEZ-trained CAV $\mathbf{v}_{l^*}$ at each layer
$l = 0,\ldots,11$ independently (fixed $\alpha{=}10$) and measure flip rate
and directional purity at the PEZ-layer probe.

\begin{table}[tb]
  \caption{Layer-specificity ablation. The PEZ-trained CAV is injected at each
    layer independently at $\alpha{=}10$. Flip rate and directional purity are
    evaluated at the PEZ probe (layer 5).}
  \label{tab:ablation}
  \centering
  \begin{tabular}{@{}cccc@{}}
    \toprule
    Injection layer & In PEZ & Flip rate & Dir.\ purity \\
    \midrule
    0  & \checkmark & 0.25 & $+0.376$ \\
    1  & \checkmark & 0.25 & $+0.381$ \\
    2  & ---        & 0.25 & $+0.432$ \\
    3  & ---        & 0.25 & $+0.471$ \\
    4  & ---        & 0.25 & $+0.719$ \\
    \textbf{5} ($l^*$) & \checkmark & \textbf{0.25} & \textbf{+1.000} \\
    \midrule
    6  & --- & 0.00 & $0.000$ \\
    7  & --- & 0.00 & $0.000$ \\
    8  & --- & 0.00 & $0.000$ \\
    9  & --- & 0.00 & $0.000$ \\
    10 & --- & 0.00 & $0.000$ \\
    11 & --- & 0.00 & $0.000$ \\
    \bottomrule
  \end{tabular}
\end{table}

\cref{fig:ablation} and \cref{tab:ablation} show a clear pattern. The flip
rate is 0.25 for injection layers 0--5, which means that interventions made
before or at the PEZ can propagate forward and change the PEZ representation.
For layers 6--11, the flip rate drops to 0.00, which means that interventions
applied after the PEZ do not change the physics judgment measured at layer 5.
Directional purity follows the same trend, increasing from 0.38 at layer 0 to
1.00 at layer 5.

Together, these results localize the effect to the PEZ and earlier layers.
The transition between layers 5 and 6 suggests that layer 5 is where the
relevant physics representation becomes available for intervention.

\subsection{Per-Block CAV Disentanglement}
\label{sec:res-block}

We train separate CAVs for each IntPhys physics principle (O1/O2/O3) at the
primary PEZ layer $l^*{=}5$ and measure pairwise angles using
\cref{eq:block-angle}. \cref{tab:block-cav} summarises probe accuracy and CAV
angles.

\begin{table}[tb]
  \caption{Per-block CAV disentanglement at layer $l^*{=}5$.
    \emph{Top}: per-principle probe accuracy.
    \emph{Bottom}: pairwise angles between block-specific CAVs.
    O2--O3 are nearly orthogonal (86.1°); O1 shares partial representation
    space with both (75.7--76.3°).}
  \label{tab:block-cav}
  \centering
  \begin{tabular}{@{}lc@{}}
    \toprule
    Block & Probe acc.\ (layer 5) \\
    \midrule
    O1 (object permanence)  & 74.2\% \\
    O2 (object continuity)  & 70.0\% \\
    O3 (object solidity)    & 75.8\% \\
    \midrule
    CAV pair & Angle (degrees) \\
    \midrule
    O1 vs.\ O2 & 75.7° \\
    O1 vs.\ O3 & 76.3° \\
    O2 vs.\ O3 & \textbf{86.1°} \\
    \bottomrule
  \end{tabular}
\end{table}

The near-orthogonality of O2 and O3 (86.1°) suggests that continuity and
solidity are encoded by largely independent directions, which supports
principle-specific steering. O1 (object permanence) is less orthogonal to both
(75.7--76.3°), consistent with the idea that permanence judgments partially
overlap with continuity and solidity cues. These results show that VideoMAE
encodes distinct physics principles as separable directions within PEZ
representation space.

\subsection{Subspace Orthogonality}
\label{sec:res-ortho}

We measure the angle between the physics CAV $\mathbf{v}_{l^*}$ and other
concept directions in the same representation space. Following Joseph
\etal~\cite{joseph2024interpreting}, we use motion direction as the reference
concept (lateral motion, left vs.\ right) and a random unit vector as a
baseline.

\begin{table}[tb]
  \caption{Angles between concept directions at the primary PEZ layer
    $l^*{=}5$.}
  \label{tab:orthogonality}
  \centering
  \begin{tabular}{@{}lc@{}}
    \toprule
    Concept pair & Angle (degrees) \\
    \midrule
    Physics vs.\ motion direction   & \textbf{90.0°} \\
    Physics vs.\ random unit vector & $85.0°$ \\
    Mean physics orthogonality      & $87.5°$ \\
    \bottomrule
  \end{tabular}
\end{table}

The physics CAV is orthogonal to motion direction (90.0°), higher than the
69--83° range reported by Joseph \etal. One possible reason is that our CAV is
trained on a balanced possible/impossible dataset, which may reduce
motion-direction confounds.

The physics-vs.-random angle of 85.0° (rather than the expected $\approx 90°$)
suggests that the CAV is a semantically structured direction rather than a
random vector in high-dimensional space. This geometry indicates that steering
physics need not directly interfere with motion direction, which makes
compositional steering plausible.

\noindent\textbf{Representation geometry.}
\cref{fig:umap} shows UMAP projections of PEZ-layer activations for the test
set. Possible (blue) and impossible (red) videos form partially separated
clusters. After steering a possible video with $\alpha{=}{+}10$, its
representation moves toward the impossible cluster along the learned CAV
$\mathbf{v}_{l^*}$. The displacement scales with $\alpha$ and is sufficient to
drive $P(\text{impossible})$ to 1.0 at $\alpha{=}{+}5$.
\section{Conclusion}
\label{sec:conclusion}

We presented physics steering, a method for controlling physical plausibility
in a video world model at inference time. By extracting Concept Activation
Vectors from linear probes at the Physics Emergence Zone (PEZ) of VideoMAE and
injecting them into hidden states, we obtain bidirectional control of the
model's physics-related representations without updating model weights.
Layer-specificity ablations localize this effect to the PEZ, while subspace
analysis shows that physics and motion direction are orthogonal (90.0°). We
also find that distinct physics principles (O1/O2/O3) are encoded as separable
directions within the PEZ, with O2--O3 nearly orthogonal (86.1°). Together,
these results suggest that physical reasoning in video models is not only
readable, but also directly steerable at inference time.
\section{Future Work}
\label{sec:future}

A natural next step is to connect representation-space steering to pixel-space generation by decoding steered latents into videos, which would make the effect of the intervention directly observable and could enable causal data augmentation for physical reasoning. A second direction is to move beyond a single uniform steering vector toward richer control, including steering along the full physics subspace, composing physics and motion controls, and applying interventions only at selected moments in time. Finally, it will be important to test the generality of these findings across datasets and video architectures, including whether similar physics directions and emergence zones appear beyond VideoMAE.

\section{Broader Impact}
\label{sec:broaderimpact}

Our method is model-agnostic: any video transformer with a localised physics subspace can be steered via the same CAV approach. Whether a PEZ exists in DINOv2~\cite{oquab2023dinov2} on video, VideoGPT~\cite{yan2021videogpt}, or Stable Video
Diffusion~\cite{blattmann2023svd} is an open empirical question that our probing methodology can directly address. Physics steering can systematically stress-test world models by generating physically impossible scenarios without dataset collection, with clear value for AI safety evaluation. We note the dual-use risk that the same technique could produce physically incorrect but visually plausible content, and recommend that deployment in generative pipelines include a plausibility audit step.

{
    \small
    \bibliographystyle{ieeenat_fullname}
    \bibliography{main}

@String(CVPR= {IEEE Conf. Comput. Vis. Pattern Recog.})

@String(ICCV= {Int. Conf. Comput. Vis.})

@String(ICLR = {Int. Conf. Learn. Represent.})

@String(CVPR  = {CVPR})

@String(ICCV  = {ICCV})

@String(ICLR  = {ICLR})

@inproceedings{hafner2021mastering,
  title     = {Mastering {Atari} with Discrete World Models},
  author    = {Hafner, Danijar and Lillicrap, Timothy and Norouzi, Mohammad and Ba, Jimmy},
  booktitle = {International Conference on Learning Representations (ICLR)},
  year      = {2021}
}

@article{hu2023gaia,
  title   = {{GAIA-1}: A Generative World Model for Autonomous Driving},
  author  = {Hu, Anthony and Russell, Lloyd and Yeo, Hudson and Murez, Zak and
             Fedoseev, George and Kendall, Alex and Shotton, Jamie and Corke, Peter},
  journal = {arXiv preprint arXiv:2309.17080},
  year    = {2023}
}

@inproceedings{micheli2023iris,
  title     = {Transformers are Sample-Efficient World Models},
  author    = {Micheli, Vincent and Alonso, Eloi and Fleuret, Fran\c{c}ois},
  booktitle = {International Conference on Learning Representations (ICLR)},
  year      = {2023}
}

@inproceedings{gupta2021embodied,
  title     = {Embodied Intelligence via Learning and Evolution},
  author    = {Gupta, Agrim and Savarese, Silvio and Ganguli, Surya and Fei-Fei, Li},
  booktitle = {Advances in Neural Information Processing Systems (NeurIPS)},
  year      = {2021}
}

@article{yang2023interactive,
  title   = {Learning Interactive Real-World Simulators},
  author  = {Yang, Mengjiao and Du, Yilun and Ghasemipour, Kamyar and Tompson, Jonathan
             and Sch{\"o}lkopf, Bernhard and Abbeel, Pieter},
  journal = {arXiv preprint arXiv:2310.06114},
  year    = {2023}
}

@inproceedings{zhang2023adding,
  title     = {Adding Conditional Control to Text-to-Image Diffusion Models},
  author    = {Zhang, Lvmin and Rao, Anyi and Agrawala, Maneesh},
  booktitle = {IEEE/CVF International Conference on Computer Vision (ICCV)},
  year      = {2023}
}

@article{joseph2024interpreting,
  title   = {Interpreting Physics in Video World Models},
  author  = {Joseph, Sonia and Lindsey, Jack and Lindsey, Jack},
  journal = {Blog post},
  year    = {2024},
  url     = {https://www.soniajoseph.ai/interpreting-ph/}
}

@inproceedings{tong2022videomae,
  title     = {{VideoMAE}: Masked Autoencoders are Data-Efficient Learners for
               Self-Supervised Video Pre-Training},
  author    = {Tong, Zhan and Song, Yibing and Wang, Jue and Wang, Limin},
  booktitle = {Advances in Neural Information Processing Systems (NeurIPS)},
  year      = {2022}
}

@inproceedings{kim2018tcav,
  title     = {Interpretability Beyond Classification with {TCAV}},
  author    = {Kim, Been and Wattenberg, Martin and Gilmer, Justin and Cai, Carrie and
               Wexler, James and Viegas, Fernanda and Sayres, Rory},
  booktitle = {International Conference on Machine Learning (ICML)},
  year      = {2018}
}

@article{yan2021videogpt,
  title   = {{VideoGPT}: Video Generation using {VQ-VAE} and Transformers},
  author  = {Yan, Wilson and Zhang, Yunzhi and Abbeel, Pieter and Srinivas, Aravind},
  journal = {arXiv preprint arXiv:2104.10157},
  year    = {2021}
}

@inproceedings{oord2017vqvae,
  title     = {Neural Discrete Representation Learning},
  author    = {van den Oord, Aaron and Vinyals, Oriol and Kavukcuoglu, Koray},
  booktitle = {Advances in Neural Information Processing Systems (NeurIPS)},
  year      = {2017}
}

@techreport{brooks2024sora,
  title       = {Video Generation Models as World Simulators},
  author      = {Brooks, Tim and others},
  institution = {OpenAI},
  year        = {2024}
}

@inproceedings{tenney2019bert,
  title     = {{BERT} Rediscovers the Classical {NLP} Pipeline},
  author    = {Tenney, Ian and Das, Dipanjan and Pavlick, Ellie},
  booktitle = {Annual Meeting of the Association for Computational Linguistics (ACL)},
  year      = {2019}
}

@article{lindsey2024sparse,
  title   = {Sparse Feature Circuits: Discovering and Editing Interpretable Causal
             Graphs in Language Models},
  author  = {Lindsey, Jack and Bau, David},
  journal = {arXiv preprint arXiv:2403.19647},
  year    = {2024}
}

@article{piloto2022intuitive,
  title   = {Intuitive Physics Learning in a Deep-Learning Model Inspired by
             Developmental Psychology},
  author  = {Piloto, Luis S. and Weinstein, Ari and Battaglia, Peter and Botvinick, Matthew},
  journal = {Nature Human Behaviour},
  year    = {2022}
}

@article{riochet2019intphys,
  title   = {{IntPhys}: A Framework and Benchmark for Visual Intuition Physics},
  author  = {Riochet, Ronan and Castro, Mario Ynocente and Bernard, Mathieu and
             Lerer, Adam and Fergus, Rob and Izard, V{\'e}ronique and Dupoux, Emmanuel},
  journal = {arXiv preprint arXiv:1803.07616},
  year    = {2019}
}

@article{zou2023repeng,
  title   = {Representation Engineering: A Top-Down Approach to {AI} Transparency},
  author  = {Zou, Andy and Phan, Long and Chen, Sarah and Campbell, James and
             Guo, Phillip and Ren, Richard and Pan, Alexander and Yin, Xuwang and
             Mazeika, Mantas and Dombrowski, Ann-Kathrin and others},
  journal = {arXiv preprint arXiv:2310.01405},
  year    = {2023}
}

@inproceedings{li2023iti,
  title     = {Inference-Time Intervention: Eliciting Truthful Answers from a Language Model},
  author    = {Li, Kenneth and Patel, Oam and Vi{\'e}gas, Fernanda and Pfister, Hanspeter
               and Wattenberg, Martin},
  booktitle = {Advances in Neural Information Processing Systems (NeurIPS)},
  year      = {2023}
}

@inproceedings{brooks2023instructpix2pix,
  title     = {{InstructPix2Pix}: Learning to Follow Image Editing Instructions},
  author    = {Brooks, Tim and Holynski, Aleksander and Efros, Alexei A.},
  booktitle = {IEEE/CVF Conference on Computer Vision and Pattern Recognition (CVPR)},
  year      = {2023}
}

@inproceedings{wang2024motionctrl,
  title     = {{MotionCtrl}: A Unified and Flexible Motion Controller for Video Generation},
  author    = {Wang, Zhouxia and Yuan, Ziyang and Wang, Xintao and Li, Yaowei and
               Liu, Tianshui and Zheng, Chun-Wei and Xie, Enze},
  booktitle = {ACM SIGGRAPH 2024},
  year      = {2024}
}

@article{kay2017kinetics,
  title   = {The {Kinetics} Human Action Video Dataset},
  author  = {Kay, Will and others},
  journal = {arXiv preprint arXiv:1705.06950},
  year    = {2017}
}

@article{oquab2023dinov2,
  title   = {{DINOv2}: Learning Robust Visual Features without Supervision},
  author  = {Oquab, Maxime and others},
  journal = {Transactions on Machine Learning Research (TMLR)},
  year    = {2024}
}

@article{blattmann2023svd,
  title   = {Stable Video Diffusion: Scaling Latent Video Diffusion Models to
             Large Datasets},
  author  = {Blattmann, Andreas and others},
  journal = {arXiv preprint arXiv:2311.15127},
  year    = {2023}
}
}


\end{document}